\documentclass[10pt,twocolumn,letterpaper]{article}

\usepackage{cvpr}              

\usepackage{graphicx}
\usepackage{amsmath}
\usepackage{amssymb}
\usepackage{booktabs}

\usepackage{caption}
\usepackage{subcaption}
\usepackage{algpseudocode}
\usepackage{lipsum}
\usepackage{multirow}
\usepackage{graphicx}
\usepackage{xcolor}
\usepackage{float}

\usepackage[pagebackref,breaklinks,colorlinks]{hyperref}

\usepackage[capitalize]{cleveref}
\crefname{section}{Sec.}{Secs.}
\Crefname{section}{Section}{Sections}
\Crefname{table}{Table}{Tables}
\crefname{table}{Tab.}{Tabs.}

\def\cvprPaperID{*****} 
\def\confName{CVPR}
\def\confYear{2023}

\begin{document}

\title{An Overview of Challenges in Egocentric Text-Video Retrieval}

\author{Burak Satar$^{1, 2}$\\
\and
Hongyuan Zhu$^{1}$\\
\and
Hanwang Zhang$^{2}$\\
\and
Joo Hwee Lim$^{1, 2}$
\and
$^{1}$Institute for Infocomm Research, A*STAR, Singapore
\and
$^{2}$School of Computer Science and Engineering, NTU, Singapore\\
{\tt\small \{burak\_satar, zhuh, joohwee\}@i2r.a-star.edu.sg}, \tt\small hanwangzhang@ntu.edu.sg}

\maketitle

\begin{abstract}
   Text-video retrieval contains various challenges, including biases coming from diverse sources. We highlight some of them supported by illustrations to open a discussion. Besides, we address one of the biases, frame length bias, with a simple method which brings a very incremental but promising increase. We conclude with future directions.
\end{abstract}

\section{Introduction}
\label{sec:intro}

Text-video retrieval aims to retrieve relevant videos given a textual query or vice versa. Recent studies \cite{chen_hgr_cvpr_2020, falcon2022learning, satar2022rome, jpose, Bain_Nagrani_Varol_Zisserman_2021, satar2022exploiting, egovlp} show significant advances either by using exocentric datasets \cite{miech19howto100m} or egocentric datasets \cite{Damen2022RESCALING, ego4d_dataset}. Many works address challenges for various text-video-related tasks \cite{Yoon_debias_vcmr_2022, yang2021deconfounded, single_frame_bias_2022, broome_bias_action_reg_wacv_2023, Hara_2021_CVPR_action_bias} including some biases that can affect their results. However, to the best of our knowledge, no work has yet to be dedicated to addressing the biases in the text-video retrieval task. 

Using a baseline work, we share insights specifically for the text-video retrieval task. The first is on biases coming from static visual features, the object's size or the camera's angle, and a possible effect from the punctuation in the textual query. The second one is ambiguity, especially in the visual features. The last one is the frame length bias and a preliminary method and experiment.
The contribution of this extended abstract is twofold. 1) We highlight various biases to bring discussion and awareness. 2) We share a baseline approach and the result for one of the biases.

\section{Challenges to Report}

We report possible challenges we face during our previous and ongoing works, especially on the Epic-Kitchen-100 dataset. These include various specific examples after examining the dataset and the possible effects of the biases on the retrieved items. While we only report the first two challenges, we suggest a baseline solution for the third.

\subsection{Textual Bias or Visual Bias?}

\begin{figure*}[!htb]
\begin{tabular}{ccc}
\fbox{\includegraphics[width=5.1cm]{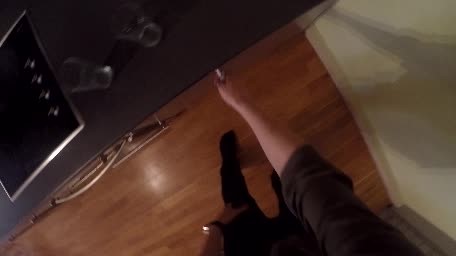}}&
\fbox{\includegraphics[width=5.1cm]{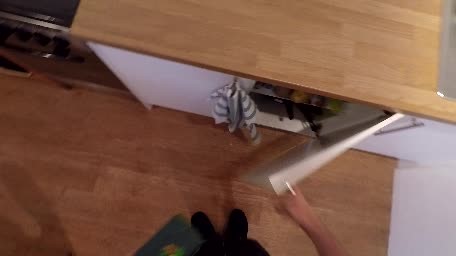}}&
\fbox{\includegraphics[width=5.1cm]{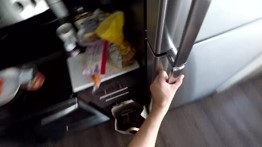}}\\
\fbox{\includegraphics[width=5.1cm]{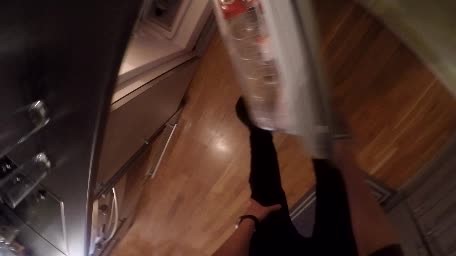}}& 
\fbox{\includegraphics[width=5.1cm]{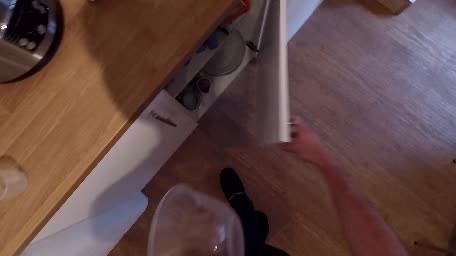}}& 
\fbox{\includegraphics[width=5.1cm]{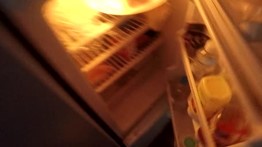}}\\
(a) caption: 'open fridge.'  %
& (b) caption: 'open cupboard' 
& (c) caption: 'opening the fridge'\\
It fails to retrieve GT clip  %
& Visual similarity with (a)
& Visual dissimilarity with (a)
\end{tabular}
\vspace{0.3cm}
\caption{ Textual queries and samples from the retrieved clips for a possible bias from the textual and visual side. While the punctuation in the first sample could bring a bias, visual discrepancy inside the 'fridge' noun class may affect the model. For instance, the fridge is small-size, only in one clip.}
\label{fig:visual_textual_bias}
\end{figure*}

In the first example, we examine a sample in the text-to-video retrieval task as shown in Figure \ref{fig:visual_textual_bias}. We find out that 'open fridge.' query, fails severely to retrieve the Ground Truth (GT) video, ranking it over 1,000. When we check the first ten retrieved videos, their corresponding captions are all about 'open cupboard'. After we examine the other queries, which include (open, fridge) as a pair of verbs and nouns, the results are pretty good. Thus, one may say that the dot '.' punctuation would bring a bias. 

While the punctuation would affect the model, another possibility may arise after checking the visual samples. For instance, only in the 'open fridge.' example, the refrigerator is small-size based on the visual frames. In contrast, the fridge is normal-size in all the other examples mentioned (open, fridge). However, why does it even matter? Because there are many visual similarities between the samples from the cupboards and the small-size fridges, such as size similarity and spatial location similarity that stay on the lower side of the kitchen, which also makes the angle of the head-mounted camera similar. Basically, the visual similarity of the small-size fridge with the cupboards and dissimilarity with the other fridges may affect the retrieved clips. 

Lastly, we may share one more observation that the variation in the verb or noun does not affect the model's performance. When the textual queries are 'opening the fridge', 'open the fridge', 'opening fridge' and 'open fridge', the results are pleasing. Nevertheless, we note that the fridges are all normal-size in these samples.

\subsection{Ambiguity at Feature Set}

\begin{figure*}[!htb]
\begin{tabular}{cccc}
\fbox{\includegraphics[width=3.7cm]{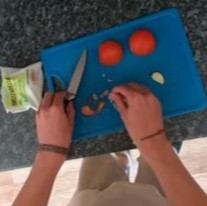}}&
\fbox{\includegraphics[width=3.7cm]{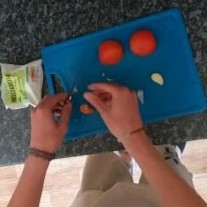}}
&
\fbox{\includegraphics[width=3.7cm]{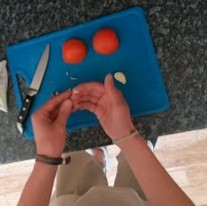}}
&
\fbox{\includegraphics[width=3.7cm]{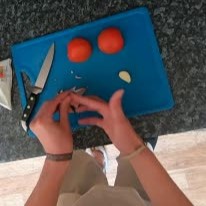}}
\\
(a) Sampled frame, 1st. %
& (b) Sampled frame, 2nd. & (c) Sampled frame, 24th. & (d) Sampled frame, 25th.
\end{tabular}
\vspace{0.3cm}
\caption{Various sampled frames from the caption: 'pick up rubbish' could bring ambiguity. For instance, the first two and last two frames may suggest the action of 'take knife' and 'clean hand'. By using the same query, our baseline method retrieves a video clip regarding 'take knife' at fourth rank.}
\label{fig:ambiguity}
\end{figure*}

Another example is that when we retrieve various clips based on a textual query, we find various clips that are completely irrelevant to the text, as shown in Figure \ref{fig:ambiguity}. While some of them may be explained by a bias, as shown in the next part, others happen due to the ambiguity in the visual feature. For instance, when checking uniformly sampled frames of the action as a pair of (verb, noun), we see that the first and last two sampled frames could be irrelevant to the action itself. Then, it also affects the result after examining the retrieved video clips. Considering a caption of 'pick up rubbish', four sampled frames out of twenty-five contain irrelevant activities to the GT caption, which could be seen as 'take knife', 'put knife back' or 'clean hand'. As a result, the model fails to rank the GT video clip and ranks it over 300. However, more importantly, a video clip with its corresponding caption 'take knife' rank at 4, showing that this ambiguity may affect the retrieved clips.

\subsection{Frame Length Discrepancy}

\begin{figure}[!htb]
    \centering
    \begin{subfigure}[b]{0.475\textwidth}
        \centering
        \includegraphics[width=0.80\linewidth]{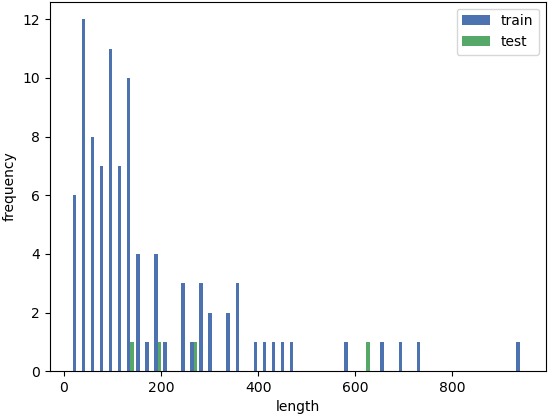}
        \caption{The histogram shows the discrepancy of a caption: \textit{'pick up rubbish'}. The GT recall is at 397th rank. The same disparity occurs among many classes (semantic pairs of verb and noun) between the train/test set.}
        \label{fig:frame_lenght_bias_a}
    \end{subfigure}
    \hfill
    \begin{subfigure}[b]{0.475\textwidth}
        \centering
        \includegraphics[width=0.80\linewidth]{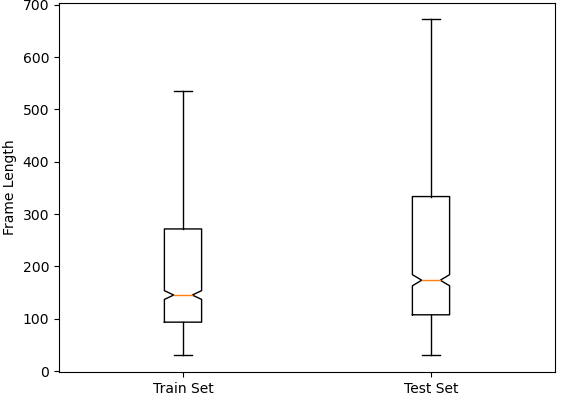}
        \caption{Average frame length comparison between train and test set showing that clips in the test set are longer than the training set,}
        \label{fig:frame_lenght_bias_b}
    \end{subfigure}
    \caption{Bias Verification on Epic-Kitchens-100. }
    \label{fig:frame_lenght_bias}
\end{figure}

We report another bias that comes from the train and test set discrepancy in the dataset, as shown in Figure \ref{fig:frame_lenght_bias_a}. The same trend can be seen after examining the classes between the train and the test set. We refer to the class as a semantic pair of a (verb, noun). Our observations show that the problem of the length bias in model training could harm retrieval performance on test sets with length discrepancy. More specifically, frame length discrepancy between training and test sets of trimmed video clips causes non-relevant video clips to be retrieved just because their frame length is similar to the training class's average frame length.

\section{Method for Frame Length Bias}

We propose a simple method to mitigate the bias by removing short and long clips of certain classes to lower the discrepancy between the train and test sets. A naive method would delete video clips only for one video class and then train the model. After repeating the same approach for classes, all the similarity matrices would be summed up. 

Another simple method would be deleting long and short video clips in a class while repeating this for every class by following two basic rules. 1) lower the discrepancy to a pre-set margin $\alpha$ 2) the number of video clips for a class cannot be less than a certain number $\beta$ to ensure having enough samples for training. Once the discrepancy gets lower, we do the training. 

\section{Experiments}

\textbf{Implementation details.} We follow a baseline method \cite{chen_hgr_cvpr_2020, satar_2021} that contains a text encoder, a video encoder and a text-video matching module. The pre-set margin $\alpha$ is chosen as 60 to reduce the discrepancy for every class in our simple method. For $\beta$, we decide that a class, meaning a semantic pair of verbs and nouns, should include more than ten samples. Specifically, we delete 2,392 clips from 164 classes, equivalent to 3.6\% of all the data.

\textbf{Dataset.} We use Epic-Kitchen-100 \cite{Damen2022RESCALING} for our experiments, which is an egocentric dataset capturing daily kitchen actions. We utilize uniformly sampled features shared by \cite{Kazakos_Nagrani_Zisserman_Damen_2019}.

\textbf{Evaluation metrics.} We use recall and mean average precision (mAP) as usual. Besides, nDCG (normalized discounted cumulative gain) is also reported by following \cite{wray2021semantic}. The details about the calculation of nDCG via relevancy matrix can be found here\cite{Damen2022RESCALING, wray2021semantic}.

\textbf{Results.}
Table \ref{tab:RmvOne} shows the effect of the first simple method on an individual sample. Although this is impractical to apply for every class, we share the results since the result also re-verify the bias. It indicates that if the gap between the training and test set regarding average frame length decreases, 1) the average frame length of the top 20 retrieved video clips increases, and 2) the ground truth value at recall improves.
On the other hand, Table \ref{tab:exp_2} shows that we slightly overpass the baseline when we follow our second simple method.

\begin{table}[!htb]
\centering
\resizebox{\columnwidth}{!}{%
\begin{tabular}{c|cc|c|c|c|}
\cline{2-6}
\textbf{}                                                                                  & \multicolumn{2}{c|}{\textbf{Avg frame length}}                         &                                          &                                                                                     &                                                                                                                \\ \cline{1-3}
\multicolumn{1}{|c|}{\textbf{Method}}                                                      & \multicolumn{1}{c|}{\textbf{Train set}}            & \textbf{Test set} & \multirow{-2}{*}{\textbf{\begin{tabular}[c]{@{}c@{}}\# of\\ samples\end{tabular}}} & \multirow{-2}{*}{\textbf{\begin{tabular}[c]{@{}c@{}}GT rank\\ recall\end{tabular}}} & \multirow{-2}{*}{\textbf{\begin{tabular}[c]{@{}c@{}}Avg frame length of \\ the top 20 retrieval\end{tabular}}} \\ \hline
\multicolumn{1}{|c|}{\begin{tabular}[c]{@{}c@{}}\textit{Baseline +} \\ Remove long clips\end{tabular}} & \multicolumn{1}{c|}{{\color[HTML]{CB0000} 36.26}}  & 287.93            & 57                                       & {\color[HTML]{3166FF} 342}                                                          & {\color[HTML]{CB0000} 167.82}                                                                                  \\ \hline
\multicolumn{1}{|c|}{Baseline}                                                             & \multicolumn{1}{c|}{{\color[HTML]{CB0000} 198.06}} & 287.93            & 88                                       & {\color[HTML]{3166FF} 148}                                                          & {\color[HTML]{CB0000} 203.59}                                                                                  \\ \hline
\multicolumn{1}{|c|}{\begin{tabular}[c]{@{}c@{}}\textit{Baseline +} \\ Remove short clips\end{tabular}} & \multicolumn{1}{c|}{{\color[HTML]{CB0000} 286.05}} & 287.93            & 57                                       & {\color[HTML]{3166FF} 58}                                                           & {\color[HTML]{CB0000} 224.22}                                                                                  \\ \hline
\end{tabular}%
}
\caption{Caption: \textit{‘put down mozzarella’} for the first naive method. Red colour refers to directly proportional, while blue is inversely proportional.}
\label{tab:RmvOne}
\end{table}

\begin{table}[!htb]
\centering
\resizebox{6cm}{!}{%
\begin{tabular}{|c|cc|}
\hline
\multirow{2}{*}{\textbf{Method}} & \multicolumn{2}{c|}{\textbf{Epic Kitchen}}                    \\ \cline{2-3} 
                                 & \multicolumn{1}{c|}{\textbf{nDCG (avg)}} & \textbf{mAP (avg)} \\ \hline
Baseline                         & \multicolumn{1}{c|}{39.15}               & 38.54              \\ \hline
Baseline + Ours                  & \multicolumn{1}{c|}{39.44}               & 38.67              \\ \hline
\end{tabular}%
}
\caption{Comparison between baseline and our second simple method. Avg refers to the average between the text-to-video and the video-to-text score of nDCG.}
\label{tab:exp_2}
\end{table}

\section{Conclusion}

We see tremendous progress in the field of text-video-related tasks. However, certain biases may still stay in the loop. We report the recent challenges we face with the motivation of bringing awareness and a discussion. Plus, we suggest a naive approach to address one of them. 
\textbf{Future work.} 1) It may be challenging to disentangle the biases coming from both the visual and textual sides. We suggest doing deep-dive work on the effect of punctuations or spatial bias, including the size of the object and the angle of the camera would shed light. 
2) To reduce ambiguity, feature purification can be applied with the guidance of the GT captions and an off-the-shelf action recognition tool.
3) We aim to address the frame length bias more effectively in our ongoing work.


{\small
\bibliographystyle{ieee_fullname}
\bibliography{egbib}

\begin{thebibliography}{10}\itemsep=-1pt

\bibitem{Bain_Nagrani_Varol_Zisserman_2021}
Max Bain, Arsha Nagrani, Gul Varol, and Andrew Zisserman.
\newblock Frozen in time: A joint video and image encoder for end-to-end
  retrieval.
\newblock In {\em 2021 IEEE/CVF International Conference on Computer Vision
  (ICCV)}, page 1708–1718, Montreal, QC, Canada, Oct 2021. IEEE.

\bibitem{broome_bias_action_reg_wacv_2023}
Sofia Broomé, Ernest Pokropek, Boyu Li, and Hedvig Kjellström.
\newblock Recur, attend or convolve? on whether temporal modeling matters for
  cross-domain robustness in action recognition.
\newblock In {\em 2023 IEEE/CVF Winter Conference on Applications of Computer
  Vision (WACV)}, pages 4188--4198, 2023.

\bibitem{chen_hgr_cvpr_2020}
Shizhe Chen, Yida Zhao, Qin Jin, and Qi Wu.
\newblock Fine-grained video-text retrieval with hierarchical graph reasoning.
\newblock In {\em 2020 IEEE/CVF Conference on Computer Vision and Pattern
  Recognition (CVPR)}, pages 10635--10644, 2020.

\bibitem{Damen2022RESCALING}
Dima Damen, Hazel Doughty, Giovanni~Maria Farinella, , Antonino Furnari, Jian
  Ma, Evangelos Kazakos, Davide Moltisanti, Jonathan Munro, Toby Perrett, Will
  Price, and Michael Wray.
\newblock Rescaling egocentric vision: Collection, pipeline and challenges for
  epic-kitchens-100.
\newblock {\em International Journal of Computer Vision (IJCV)}, 130:33–55,
  2022.

\bibitem{falcon2022learning}
Alex Falcon, Giuseppe Serra, and Oswald Lanz.
\newblock Learning video retrieval models with relevance-aware online mining.
\newblock In {\em International Conference on Image Analysis and Processing},
  pages 182--194. Springer, 2022.

\bibitem{ego4d_dataset}
Kristen Grauman, Andrew Westbury, Eugene Byrne, Zachary Chavis, Antonino
  Furnari, Rohit Girdhar, Jackson Hamburger, Hao Jiang, Miao Liu, Xingyu Liu,
  Miguel Martin, Tushar Nagarajan, Ilija Radosavovic, Santhosh~Kumar
  Ramakrishnan, Fiona Ryan, Jayant Sharma, Michael Wray, Mengmeng Xu,
  Eric~Zhongcong Xu, Chen Zhao, Siddhant Bansal, Dhruv Batra, Vincent
  Cartillier, Sean Crane, Tien Do, Morrie Doulaty, Akshay Erapalli, Christoph
  Feichtenhofer, Adriano Fragomeni, Qichen Fu, Abrham Gebreselasie, Cristina
  Gonzalez, James Hillis, Xuhua Huang, Yifei Huang, Wenqi Jia, Weslie Khoo,
  Jachym Kolar, Satwik Kottur, Anurag Kumar, Federico Landini, Chao Li, Yanghao
  Li, Zhenqiang Li, Karttikeya Mangalam, Raghava Modhugu, Jonathan Munro,
  Tullie Murrell, Takumi Nishiyasu, Will Price, Paola~Ruiz Puentes, Merey
  Ramazanova, Leda Sari, Kiran Somasundaram, Audrey Southerland, Yusuke Sugano,
  Ruijie Tao, Minh Vo, Yuchen Wang, Xindi Wu, Takuma Yagi, Ziwei Zhao, Yunyi
  Zhu, Pablo Arbelaez, David Crandall, Dima Damen, Giovanni~Maria Farinella,
  Christian Fuegen, Bernard Ghanem, Vamsi~Krishna Ithapu, C.~V. Jawahar,
  Hanbyul Joo, Kris Kitani, Haizhou Li, Richard Newcombe, Aude Oliva, Hyun~Soo
  Park, James~M. Rehg, Yoichi Sato, Jianbo Shi, Mike~Zheng Shou, Antonio
  Torralba, Lorenzo Torresani, Mingfei Yan, and Jitendra Malik.
\newblock Ego4d: Around the world in 3,000 hours of egocentric video.
\newblock In {\em 2022 IEEE/CVF Conference on Computer Vision and Pattern
  Recognition (CVPR)}, page 18973–18990, New Orleans, LA, USA, Jun 2022.
  IEEE.

\bibitem{Hara_2021_CVPR_action_bias}
Kensho Hara, Yuchi Ishikawa, and Hirokatsu Kataoka.
\newblock Rethinking training data for mitigating representation biases in
  action recognition.
\newblock In {\em Proceedings of the IEEE/CVF Conference on Computer Vision and
  Pattern Recognition (CVPR) Workshops}, pages 3349--3353, June 2021.

\bibitem{Kazakos_Nagrani_Zisserman_Damen_2019}
Evangelos Kazakos, Arsha Nagrani, Andrew Zisserman, and Dima Damen.
\newblock Epic-fusion: Audio-visual temporal binding for egocentric action
  recognition.
\newblock (arXiv:1908.08498), Aug 2019.
\newblock arXiv:1908.08498 [cs].

\bibitem{single_frame_bias_2022}
Jie Lei, Tamara~L. Berg, and Mohit Bansal.
\newblock Revealing single frame bias for video-and-language learning, 2022.

\bibitem{egovlp}
Kevin~Qinghong Lin, Alex~Jinpeng Wang, Mattia Soldan, Michael Wray, Rui Yan,
  Eric~Zhongcong Xu, Difei Gao, Rongcheng Tu, Wenzhe Zhao, Weijie Kong,
  Chengfei Cai, Hongfa Wang, Dima Damen, Bernard Ghanem, Wei Liu, and
  Mike~Zheng Shou.
\newblock Egocentric video-language pretraining.
\newblock In {\em NeurIPS}. arXiv, Oct 2022.
\newblock arXiv:2206.01670 [cs].

\bibitem{miech19howto100m}
A. Miech, D. Zhukov, J.-B. Alayrac, M. Tapaswi, I. Laptev, and J. Sivic.
\newblock Howto100m: Learning a text-video embedding by watching hundred
  million narrated video clips.
\newblock In {\em ICCV}, 2019.

\bibitem{satar_2021}
Burak Satar, Zhu Hongyuan, Xavier Bresson, and Joo~Hwee Lim.
\newblock Semantic role aware correlation transformer for text to video
  retrieval.
\newblock In {\em 2021 IEEE International Conference on Image Processing
  (ICIP)}, pages 1334--1338, 2021.

\bibitem{satar2022exploiting}
Burak Satar, Hongyuan Zhu, Hanwang Zhang, and Joo~Hwee Lim.
\newblock Exploiting semantic role contextualized video features for
  multi-instance text-video retrieval epic-kitchens-100 multi-instance
  retrieval challenge 2022, 2022.

\bibitem{satar2022rome}
Burak Satar, Hongyuan Zhu, Hanwang Zhang, and Joo~Hwee Lim.
\newblock Rome: Role-aware mixture-of-expert transformer for text-to-video
  retrieval, 2022.

\bibitem{jpose}
Michael Wray, Gabriela Csurka, Diane Larlus, and Dima Damen.
\newblock Fine-grained action retrieval through multiple parts-of-speech
  embeddings.
\newblock In {\em 2019 IEEE/CVF International Conference on Computer Vision
  (ICCV)}, page 450–459, Seoul, Korea (South), Oct 2019. IEEE.

\bibitem{wray2021semantic}
Michael Wray, Hazel Doughty, and Dima Damen.
\newblock On semantic similarity in video retrieval.
\newblock In {\em CVPR}, 2021.

\bibitem{yang2021deconfounded}
Xun Yang, Fuli Feng, Wei Ji, Meng Wang, and Tat-Seng Chua.
\newblock Deconfounded video moment retrieval with causal intervention.
\newblock In {\em Proceedings of the 44th International ACM SIGIR Conference on
  Research and Development in Information Retrieval}, SIGIR '21, page 1–10.
  Association for Computing Machinery, 2021.

\bibitem{Yoon_debias_vcmr_2022}
Sunjae Yoon, Ji~Woo Hong, Eunseop Yoon, Dahyun Kim, Junyeong Kim, Hee~Suk Yoon,
  and Chang~D. Yoo.
\newblock Selective query-guided debiasing for video corpus moment retrieval.
\newblock In Shai Avidan, Gabriel Brostow, Moustapha Cissé, Giovanni~Maria
  Farinella, and Tal Hassner, editors, {\em Computer Vision – ECCV 2022},
  page 185–200, Cham, 2022. Springer Nature Switzerland.

\end{thebibliography}
}

\end{document}